\documentclass[letterpaper, 10 pt, conference]{ieeeconf}  

\IEEEoverridecommandlockouts                              

\usepackage{multirow} 
\usepackage{graphicx}      
\usepackage{amsmath}        
\usepackage{amssymb}        
\usepackage{xcolor}
\usepackage{colortbl}
\usepackage{cite}
\usepackage{booktabs} 
\usepackage{tabularx} 
\newcolumntype{C}{>{\centering\arraybackslash}X}
\usepackage{float}
\usepackage{stfloats}
\title{\LARGE \bf
	SPARK Hand: Scooping-Pinching Adaptive Robotic Hand with Kempe Mechanism for Vertical Passive Grasp in Environmental Constraints
}

\author{Jiaqi Yin, Tianyi Bi, Wenzeng Zhang, \textit{Member, IEEE}%
	\thanks{*This research is supported by the Foundation of Enhanced Student Research Training (E-SRT) and Open Research for Innovation Challenges (ORIC), X-Institute.}%
	\thanks{Jiaqi Yin is with School of Future Technology,
		Harbin Institute of Technology, Harbin, China and Laboratory of Robotics, X-Institute, Shenzhen, China
        (e-mail: yjqhit@gmail.com).}%
	\thanks{Tianyi Bi is with Shude College,
		Southern University of Science and Technology and Laboratory of Robotics, X-Institute, Shenzhen, China
        (e-mail: 12311620@mail.sustech.edu.cn).}%
	\thanks{Wenzeng Zhang is with the Laboratory of Robotics, X-Institute, Shenzhen, China, and the Department of Mechanical Engineering, Tsinghua University, Beijing, China.
        He is the corresponding author (e-mail: zhangwenzeng@x-institute.edu.cn).}
}

\begin{document}

	\maketitle
	\thispagestyle{empty}
	\pagestyle{empty}

	\begin{abstract}
		
        This paper presents the SPARK finger, an innovative passive adaptive robotic finger capable of executing both parallel pinching and scooping grasps. The SPARK finger incorporates a multi-link mechanism with Kempe linkages to achieve a vertical linear fingertip trajectory. Furthermore, a parallelogram linkage ensures the fingertip maintains a fixed orientation relative to the base, facilitating precise and stable manipulation. By integrating these mechanisms with elastic elements, the design enables effective interaction with surfaces, such as tabletops, to handle challenging objects. The finger employs a passive switching mechanism that facilitates seamless transitions between pinching and scooping modes, adapting automatically to various object shapes and environmental constraints without additional actuators. To demonstrate its versatility, the SPARK Hand, equipped with two SPARK fingers, has been developed. This system exhibits enhanced grasping performance and stability for objects of diverse sizes and shapes, particularly thin and flat objects that are traditionally challenging for conventional grippers. Experimental results validate the effectiveness of the SPARK design, highlighting its potential for robotic manipulation in constrained and dynamic environments.

	\end{abstract}

	\section{INTRODUCTION}
	
	Grasping is a fundamental capability of robotic manipulators, enabling interaction with and manipulation of objects in diverse environments. Robotic hands, serving as end-effectors, play a crucial role in various applications, ranging from industrial automation to service robotics, prosthetics, and space exploration. Despite decades of research and development, achieving reliable and adaptive grasping across objects with varying shapes, sizes, and environmental constraints remains a significant challenge. Particularly, handling thin and flat objects, such as cards or coins lying on a surface, poses unique difficulties due to limited contact areas, high precision requirements, and environmental interactions.
	
    Traditional robotic grippers, such as industrial parallel grippers, are widely utilized due to their simplicity, robustness, and reliability. However, these grippers are often limited to grasping objects of specific geometries and lack the adaptability required to handle irregular or thin objects. Over the years, robotic grasping technologies have advanced with the development of dexterous hands and soft robotic hands. Dexterous hands, inspired by human hand functionality, offer high flexibility through multiple degrees of freedom (DOFs). Notable examples include the Stanford/JPL Hand \cite{salisbury1982articulated}, the Utah/MIT Hand \cite{jacobsen1986design}, the Robonaut Hand\cite{lovchik1999robonaut}, and the DLR Hand \cite{wei2005fpga}, which have demonstrated advanced hardware designs and control strategies. However, such designs often suffer from high complexity, limited grasping force, and high costs, making them less practical for constrained industrial environments. Conversely, soft robotic hands leverage material compliance to conform to objects with diverse shapes \cite{brown2010universal, bauer2014soft}, but face challenges in controllability, precision, and robustness when interacting with thin or rigid objects.
	
	To address the trade-offs between adaptability and simplicity, researchers have explored underactuated robotic hands, which employ fewer actuators than their total degrees of freedom \cite{laliberte2002underactuation}. These designs rely on passive elements, such as springs and mechanical linkages, to achieve adaptive grasping. Representative examples include the Barrett Hand \cite{townsend2000barrett}, the RBO Hand \cite{deimel2016novel} and the LPCSA Hand \cite{chen2024novel}, which have demonstrated effective compliance and grasping versatility. However, underactuated hands often struggle in constrained environments or when manipulating thin objects lying on flat surfaces, as their passive compliance may not provide sufficient stability or precision for such tasks.
    
	\begin{figure}[!b] 
    \vspace*{-0.5cm}
        \centering 
        \includegraphics[width=0.9\columnwidth]{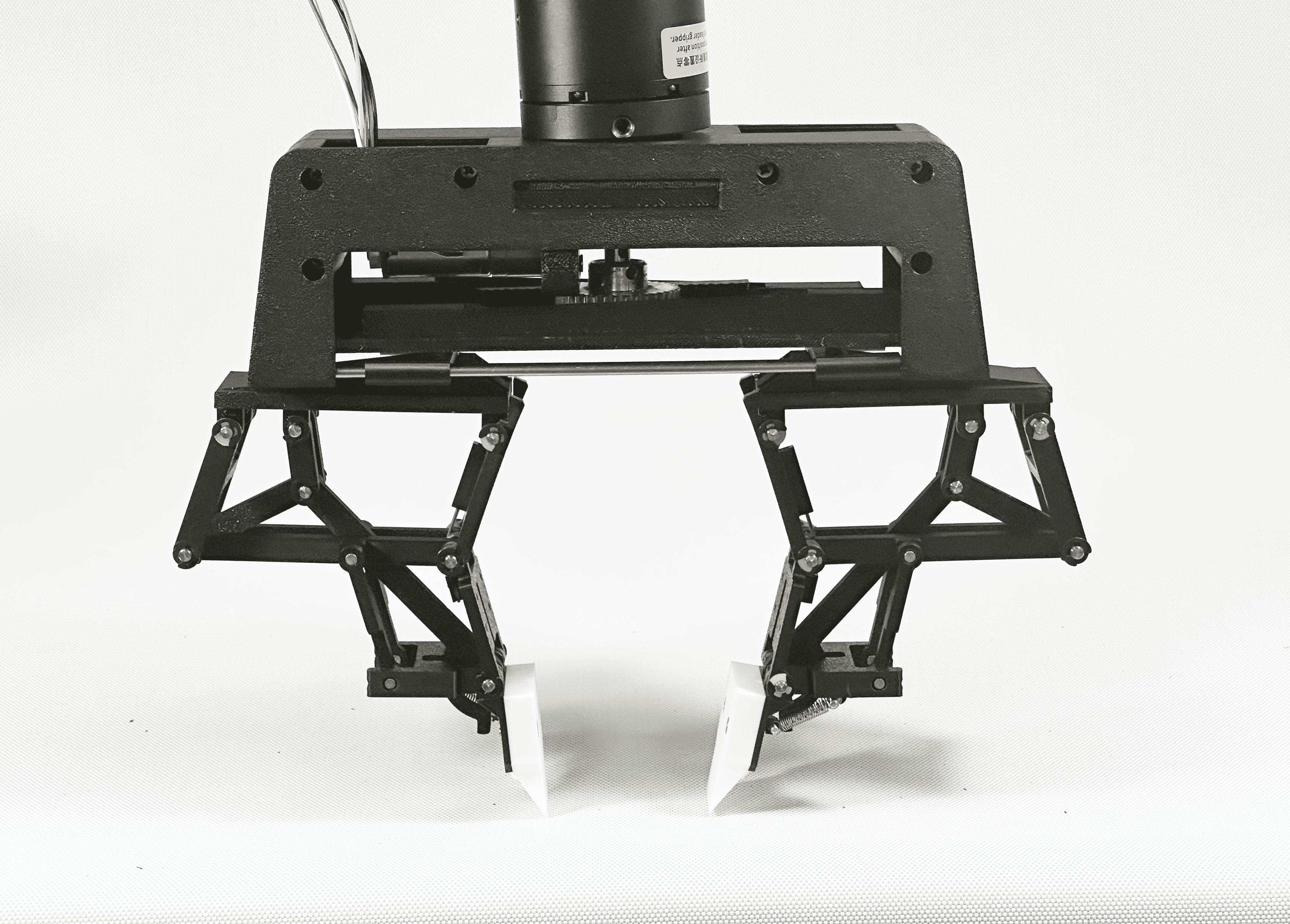} \vspace*{-0.3cm}
        \caption{SPARK Hand} 
        \label{fig:devices} 
    \end{figure}
    
	Human hands exhibit a remarkable ability to adaptively grasp objects by dynamically adjusting strategies based on object properties and environmental interactions. For instance, when grasping thin objects on a surface, humans often employ pinching motions or slide their fingertips under the object in a scooping motion. Emulating such behaviors in robotic hands has the potential to significantly enhance grasping performance in challenging scenarios. Recent innovations, such as the Omega Hand \cite{yoon2022fullypassive}, have demonstrated the feasibility of incorporating passive mechanisms to replicate human-inspired grasping strategies. These works highlight the importance of integrating environmental interactions into robotic hand designs to achieve robust and versatile grasping.

    In this paper, we introduce the SPARK Hand, illustrated in Fig.\ref{fig:devices}, a novel robotic hand featuring two SPARK fingers that enable both parallel pinching and scooping grasps.

    
    The SPARK finger incorporates a multi-link mechanism with a passive switching mechanism, facilitating automatic transitions between grasping modes based on environmental interactions. This innovative design utilizes Kempe linkages to achieve a vertical linear fingertip trajectory \cite{kempe1877}\cite{dijksman1975}, while a parallelogram linkage ensures the fingertip maintains a fixed orientation relative to the base. These features, in conjunction with elastic elements, enable the finger to adapt to various object shapes and environmental constraints, including thin and flat objects. By employing passive mechanics, the mechanism eliminates the necessity for additional actuators, resulting in a cost-effective and simplified design.
	
	
	The subsequent sections of this paper are structured as follows: Section II presents the methodology, detailing the mechanical design, theoretical analysis, and operation principles of the proposed SPARK Hand. Section III describes the experimental setup and evaluates the SPARK Hand's performance in various grasping tasks. Section IV concludes the paper by summarizing the key contributions and discussing potential avenues for future research.



\section{Design of SPARK Hand}

This section provides a comprehensive overview of the mechanical design principles and operational modes of the SPARK robotic finger mechanism. The SPARK finger supports two grasping modes, pinching and scooping, enabling adaptability to environmental constraints and diverse object geometries.

\subsection{Mechanical Design}

The SPARK robotic hand comprises a finger base and two SPARK fingers. The finger base incorporates a single actuator that provides linear motion, enabling the two fingers to move toward the opposing finger. A linear actuator was selected in preference to a rotary actuator due to its ability to maintain the configuration of the fingers relative to the global coordinate frame. Irrespective of the fingers' translational position, the linear actuator ensures uniform compliance, rendering it more suitable for interaction with static and predefined environmental constraints. Furthermore, linear actuators are extensively employed in industrial applications, enhancing compatibility with commercially available grippers and improving task adaptability.

\begin{figure}[h]
\vspace*{-0.4cm}
    \centering
    \includegraphics[width=0.7\linewidth]{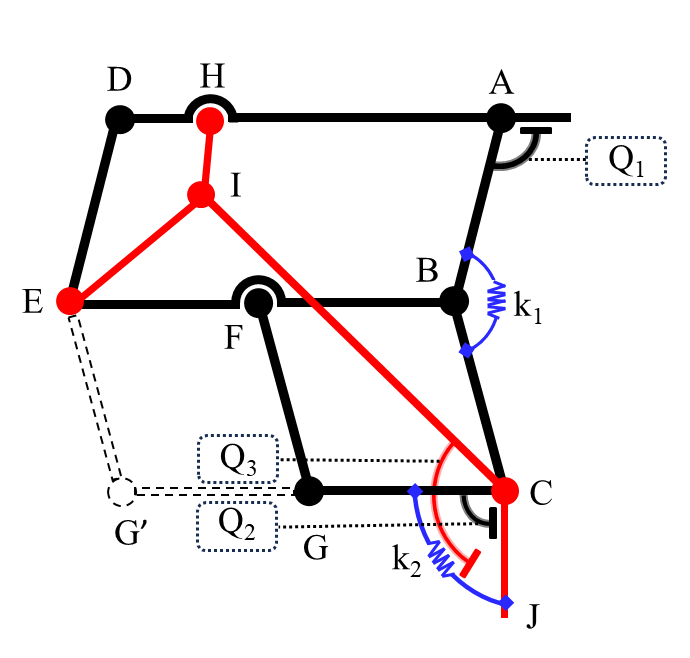}\vspace*{-0.3cm}
    \caption{Mechanism principle of the SPARK finger.}
    \label{fig:kempe_linkage}
\end{figure}

The SPARK finger's mechanism principle, depicted in Fig.\ref{fig:kempe_linkage}, illustrates its primary components: finger segments, a Kempe linkage, springs \(k_1\) and \(k_2\), and stoppers \(Q_1\), \(Q_2\), and \(Q_3\). The finger comprises three segments: proximal (\(AB\)), intermediate (\(BC\)), and distal (\(CJ\)) phalanges, interconnected by passive revolute joints. The Kempe linkage mechanism forms the finger's core structure, while springs provide compliance and restoring forces. This combination of Kempe linkage, springs, and stoppers enables the finger’s adaptive and passive switching mechanism, facilitating environmental constraint conformity and automatic transition between grasping modes.

The SPARK finger mechanism is designed with a vertical fingertip trajectory to achieve effective environmental compliance, given that the base employs a parallel gripping motion. This design ensures alignment of the pinching force with the linear actuation force, resulting in consistent grasping performance. The vertical trajectory also distributes force uniformly across the contact surface, preventing damage to fragile objects. Additionally, the gripper can grasp small and thin objects before the actuator reaches its stroke limit, enhancing operational efficiency.

\subsection{Implementation of Vertical Compliance: Kempe's Linkage}
To achieve reliable vertical compliance, a straight-line mechanism based on the Kempe linkage was implemented. This section details the adaptations made to the Kempe linkage for robotic finger applications. The linkage comprises two parallelograms and a Y-shaped linkage. The following modifications were implemented to adapt the linkage mechanism for robotic finger applications:

\begin{enumerate}
    \item The lengths of $EG'$ and $CG'$ were shortened to $FG$ and $CG$, respectively, to enhance compactness without compromising vertical trajectory performance. This design supports asymmetric scooping operations, conserves space, and accommodates scenarios where the robotic hand tilts downward to contact a surface, providing sufficient clearance.
    \item Additional springs $k_1$ and $k_2$, along with mechanical stoppers $Q_1$, $Q_2$, and $Q_3$, were integrated to enable mode switching and compliance.
    \item The base linkage $AD$ was adjusted to facilitate connection to the finger segments.
\end{enumerate}

Fig.\ref{fig:kempe_linkage} illustrates the SPARK finger mechanism based on the modified Kempe linkage. The geometric constraints that the Kempe linkage must satisfy are as follows:

The link lengths satisfy the following relationships:
$$
\scalebox{0.8}{
\begin{minipage}{\linewidth}
\begin{equation}
\begin{aligned}
L_1 &= AD = BE = CG' = CI, \\
L_2 &= AB = BC = DE = FG' = EI, \\
L_3 &= HI = DH,
\end{aligned}
\label{eq:link_lengths}
\end{equation}\end{minipage}
}
$$

$$\scalebox{0.8}{ \begin{minipage}{\linewidth} \begin{equation}
L_1 : L_2 : L_3 = 4 : 2 : 1
\label{eq:link_ratios}
\end{equation}\end{minipage} } $$

The link lengths in \eqref{eq:link_lengths} were selected to satisfy the geometric constraints of the Kempe linkage. The 4:2:1 ratio in \eqref{eq:link_ratios} ensures a consistent vertical trajectory while maintaining compactness.

The design parameters of the SPARK finger are defined as follows:

\begin{table}[ht]
\centering
\renewcommand{\arraystretch}{1.5} 
\setlength{\tabcolsep}{8pt} 

\caption{Design Parameters of SPARK finger}
\label{tab:design_parameters}

\resizebox{\linewidth}{!}{\begin{tabular}{|c|c|c|c|c|c|}
\hline
\rowcolor{gray!20} \textbf{$L_1$} & \textbf{$L_2$} & \textbf{$L_3$} & \textbf{$CJ$} & \textbf{$CG$} & \textbf{$FG$} \\ \hline
80 mm          & 40 mm          & 20 mm          & 28.8 mm       & 40 mm         & 40 mm         \\ \hline
\rowcolor{gray!20} \textbf{$\Delta h_1$} & \textbf{$\Delta h_2$} & \textbf{$\Delta \theta_{C1}$} & \textbf{$Q_1$} & \textbf{$Q_2$} & \textbf{$Q_3$} \\ \hline
15.8 mm         & 14.6 mm        & 22.8°          & 113.2°            & 90°           & 83.0°        \\ \hline
\end{tabular}}

\end{table}

\subsection{Operation Principles}

The proposed finger mechanism operates in two distinct modes: pinching and scooping. The working principles of these modes are elucidated as follows:

\subsubsection{Pinching Mode}

In this mode, the finger executes a linear clamping motion to securely grasp objects. Initially, the finger descends until the distal segment \(CJ\) establishes contact with the target surface. Stopper \(Q_2\) ensures the distal segment maintains its straight configuration, enabling precise pinching motions. This mode demonstrates particular efficacy in grasping regular objects with flat surfaces, such as cubes and cylinders.

\subsubsection{Scooping Mode}

When the finger continues its downward movement, overcoming the compressive force of torsion spring \(k_1\), and the downward displacement reaches \(\Delta h_1\), stopper \(Q_3\) contacts the distal segment \(CJ\). Further downward movement by \(\Delta h_2\) overcomes the compressive force of tension spring \(k_2\), causing the distal segment \(CJ\) to rotate inward by an angle of \(\Delta \theta_{C1}\). This creates a scoop-like configuration, enabling the finger to slide under thin or flat objects. The transition to scooping mode occurs passively without requiring additional actuators or sensors.

By adjusting the finger's tilt angle, the mechanism can perform asymmetric scooping operations. In this scenario, one finger performs scooping while the other remains vertical, allowing for the pinching of wide and thin objects such as cards.

\begin{figure}[h]
    \centering
    \includegraphics[width=1\linewidth]{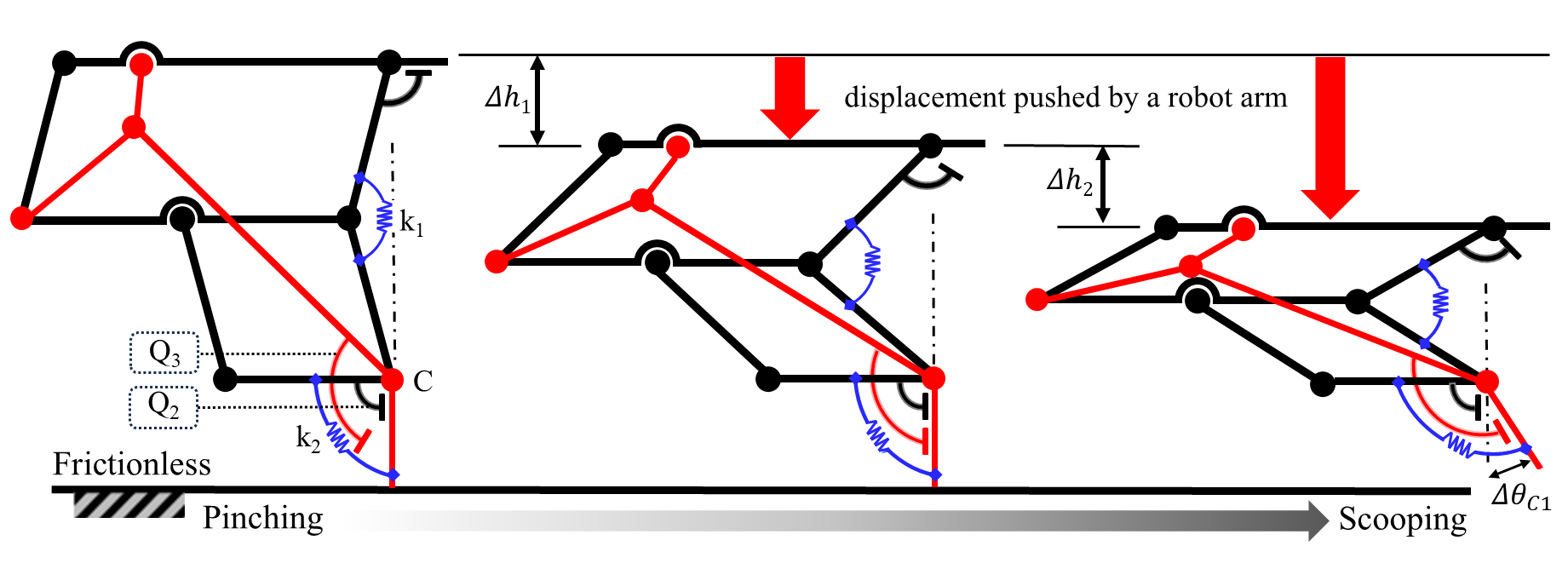}
    \caption{Fingertip transition from pinching to scooping poses using stoppers.}
    \label{fig:2}
\end{figure}

\begin{figure}[!b]
\vspace*{-0.6cm}
        \centering
        \includegraphics[width=0.57\linewidth]{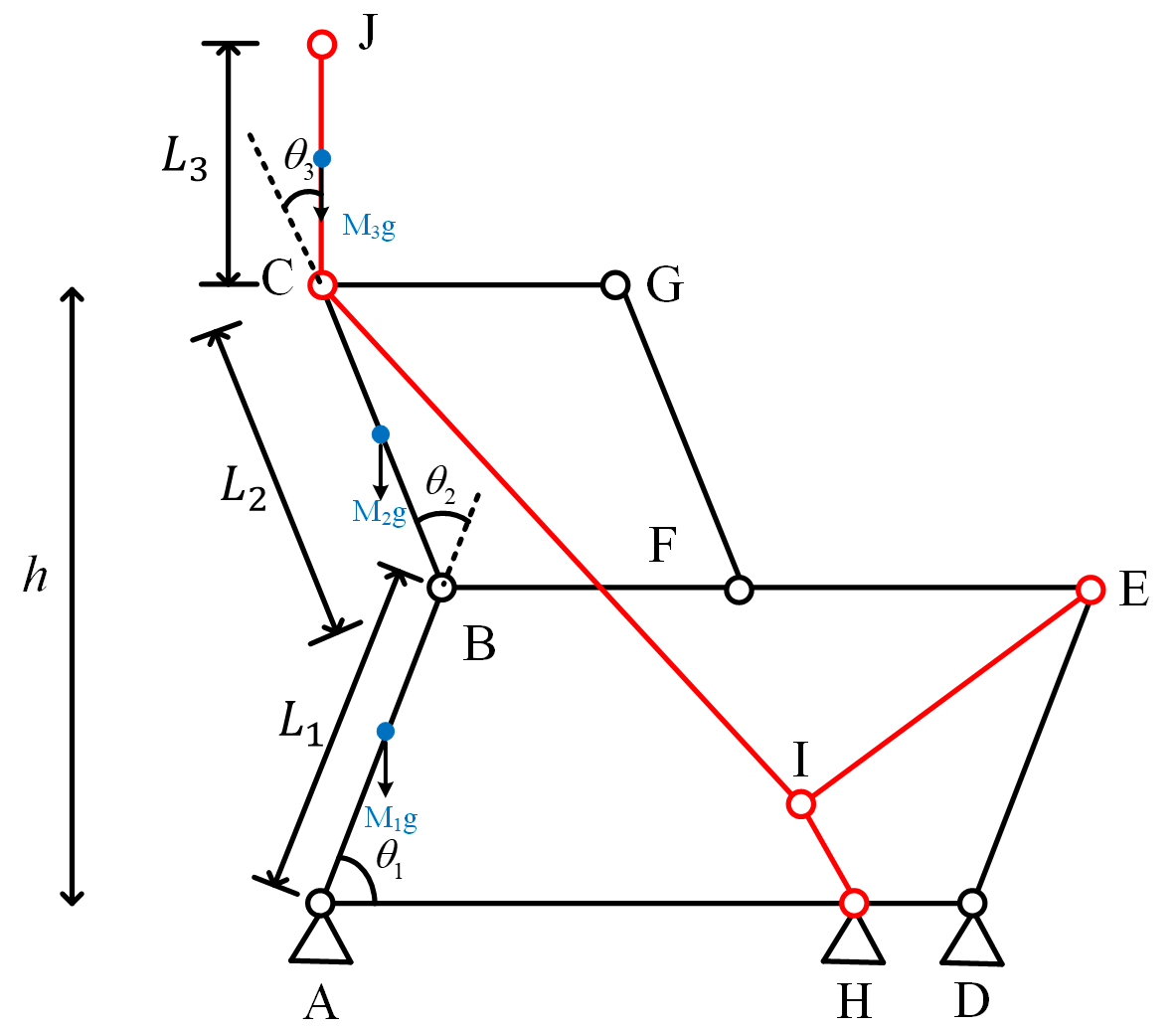}\vspace*{-0.3cm}
        \caption{Mathematical model.}
        \label{Mathematical model}
    \end{figure}
    
Fig.\ref{fig:2} illustrates the transition process of the SPARK finger from the pinching pose to the scooping pose. During the finger's pinching motion, the first and second segments exhibit adaptability to the shape and size of the grasped object due to the effect of \(k_1\).


        \section{Analysis of the SPARK finger}
        To evaluate the performance characteristics of the
    SPARK Hand, kinematic analysis and grasping
    force analysis were conducted on a single SPARK finger.

        \subsection{Kinematic Analysis of the SPARK Hand}

    A mathematical model of the robotic finger was established by representing the axle as a point and the rod as a line, as illustrated in Fig.\ref{Mathematical model}.

    The reference coordinate system was established with each rotational joint of the finger as the origin. The D-H parameters for the mathematical model of the finger were derived as follows:

    \begin{table}[ht]
        \centering
        \caption{Standard D-H Parameters for SPARK finger}
        \begin{tabularx}{\columnwidth}{CCCCC}
          \toprule
         Link & \(a_i\) & \(\alpha_i\) & \(d_i\) & \(\theta_i\) \\  \midrule
          1 & \(L_1\) & 0 & 0 & \(\theta_1\) \\ 
          2 & \(L_2\) & 0 & 0 & \(\theta_2\) \\ 
          3 & \(L_3\) & 0 & 0 & \(\theta_3\) \\  \bottomrule
        \end{tabularx}
        \label{tab1}
    \end{table}

    From the D-H table, the transformation matrix is derived:

    $$\scalebox{0.8}{ \begin{minipage}{\linewidth} \begin{equation}
    A_i = \begin{bmatrix}
    \cos \theta_i & -\sin \theta_i & 0 & L_i \cos \theta_i \\
    \sin \theta_i & \cos \theta_i & 0 & L_i \sin \theta_i \\
    0 & 0 & 1 & 0 \\
    0 & 0 & 0 & 1
    \end{bmatrix}
    \end{equation}\end{minipage} } $$

    The transformation matriices for the three rods are expressed as:
    $$\scalebox{0.8}{ \begin{minipage}{\linewidth} \begin{equation}
        T_1^0 = A_1, \quad T_2^0 = A_1 A_2, \quad T_3^0 = A_1 A_2 A_3
    \end{equation}\end{minipage} } $$
    
    Introduction of the Jacobi Matrix is expressed as:
    $$\scalebox{0.8}{ \begin{minipage}{\linewidth} \begin{equation}
        \xi = J \dot{\theta}, \quad \xi = \begin{bmatrix} V_n^0 \\ W_n^0 \end{bmatrix}, \quad J = \begin{bmatrix} J_v \\ J_w \end{bmatrix}
    \end{equation}\end{minipage} } $$
    
    $$\scalebox{0.8}{ \begin{minipage}{\linewidth} \begin{equation}
        J_{vi} = Z_{i-1}^0 \times (O_n - O_{i-1}), \quad J_{wi} = Z_{i-1}^0
    \end{equation}\end{minipage} } $$

\begin{figure}[ht]
    \centering
    \includegraphics[width=1\linewidth]{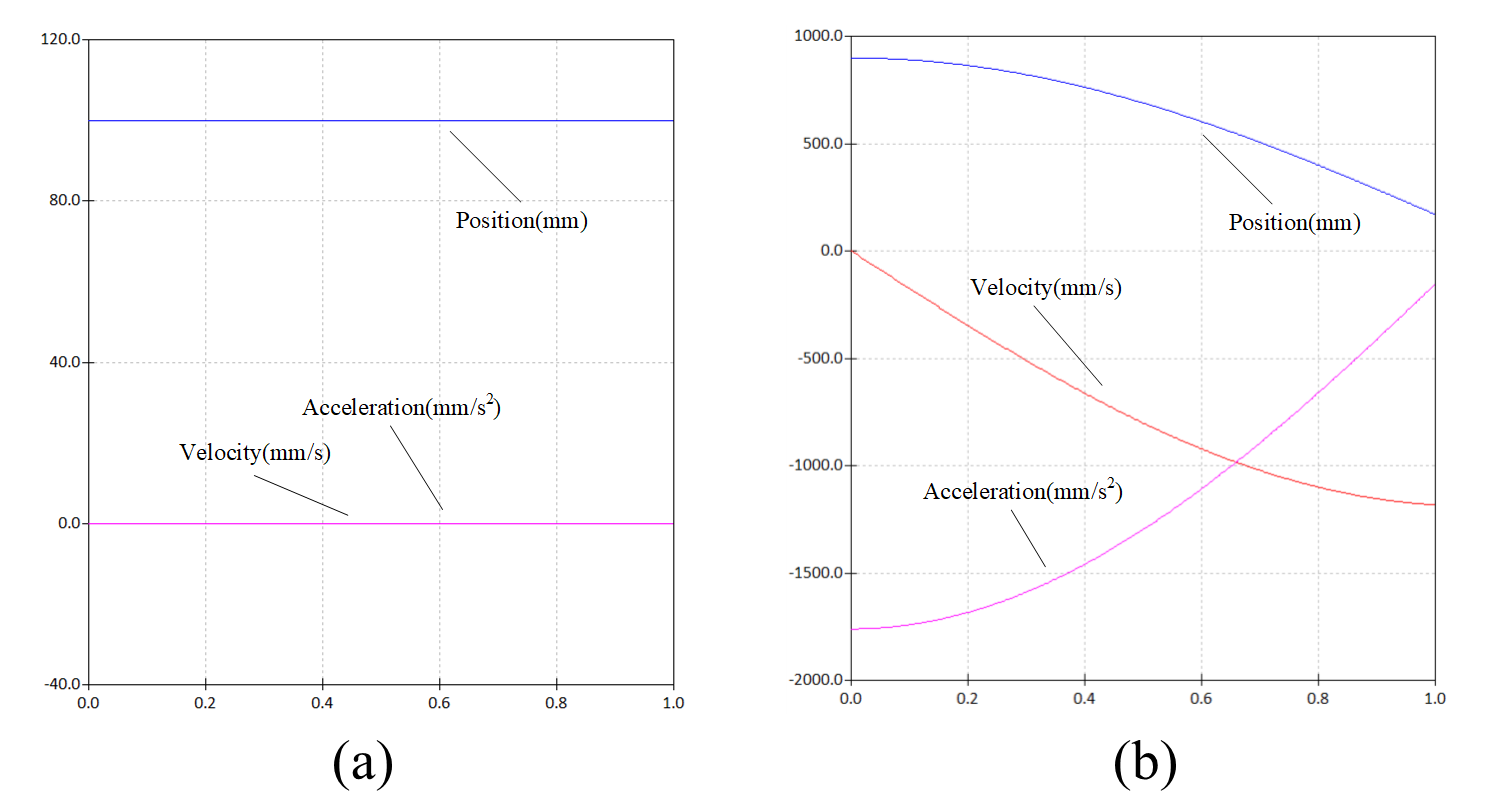}
    \caption{Motion characteristic of the SPARK Hand.(a) Horizontal kinematic characteristic of the mechanism;(b) Vertical kinematic characteristic of the mechanism.}
    \label{fig:SAM}
\end{figure}

    According to the derived equations and in conjunction with kinematic simulation software, when the mechanism is activated, the trajectory of the fingertip can be visualized, as shown in Fig.\ref{fig:SAM} .

    The motion characteristics of the fingertip, when driven by a uniform velocity, are illustrated in Fig.\ref{fig:SAM}. The analysis focused primarily on the fingertip's motion, revealing smooth movements except at initiation in the x-direction, which is ideal for handling fragile items such as eggs. In the y-direction, the displacement remains consistently zero, confirming the linear motion.

    Combining the aforementioned kinematical equations using the Lagrangian method, the kinetic equations were derived. The Jacobi matrix was subsequently obtained to determine the velocity and angular velocity of the finger during motion, as well as the kinetic energy of the rigid body:
        
    $$\scalebox{0.8}{ \begin{minipage}{\linewidth} \begin{equation}
    K = \frac{1}{2} m v^T v + \frac{1}{2} \omega^T \omega
    \end{equation}\end{minipage} } $$
    
    $$\scalebox{0.8}{ \begin{minipage}{\linewidth} \begin{equation}
    \begin{split}
      P &= \sum_{i=1}^{3} P_i\\
        &= m_1 g L_{c1} \sin(\theta_1) \\
        &\quad + m_2 g \big( L_1 \sin(\theta_1) + L_{c2} \sin(\theta_1 + \theta_2) \big) \\
        &\quad + m_3 g \big( L_1 \sin(\theta_1) + L_2 \sin(\theta_1 + \theta_2) \\
        &\quad + L_{c3} \sin(\theta_1 + \theta_2 + \theta_3) \big)
    \end{split}
    \end{equation}\end{minipage} } $$

    Then substituting into the robot dynamics equations:
    
    $$\scalebox{0.8}{ \begin{minipage}{\linewidth} \begin{equation}
    M(\theta) \ddot{\theta} + C(\theta, \dot{\theta}) \dot{\theta} + G(\theta) = \tau
    \end{equation}\end{minipage} } $$

    In practical robotic hand design, the results of these equations serve as a reference for mechanism design and optimization.

        \subsection{Grasping Force Analysis of the SPARK Hand}

\begin{figure}[ht]
    \centering
    \includegraphics[width=1\linewidth]{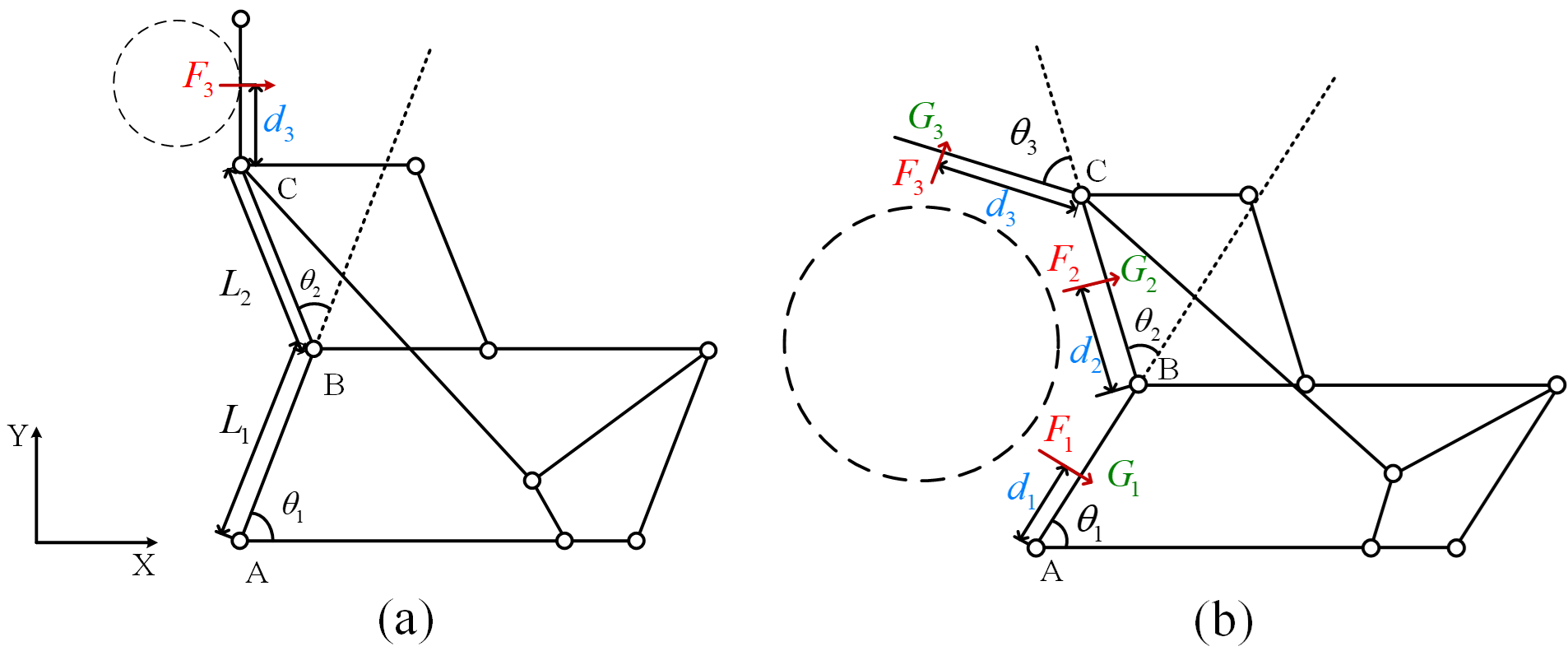}
    \caption{Hydrostatic model. (a) linear-parallel pinching force
analysis; (b) self-adaptive grasping force analysis.}
    \label{fig:force}
    \end{figure}

	In addition to the Kinematic Analysis, this subsection elaborates on the static mechanics of the SPARK finger. A static model was established based on the finger structure, with the corresponding physical meanings of the symbols depicted in Fig \ref{fig:force} and the accompanying table.

    \begin{table}[h]
    \centering
    \caption{Physical Symbols and Their Meanings}
    \begin{tabular}{cp{5cm}c}
    \toprule
    \textbf{Quantity} & \textbf{Notation} & \textbf{Unit} \\
     \midrule
   \multirow{2}{*}{$F_1, F_2, F_3$} & Grasping power of the first, second and third phalanxes. & N \\
    
    \multirow{2}{*}{$L_1, L_2, L_3$} & Length of the first, second and third phalanxes. & mm \\
    
   \multirow{3}{*}{$d_1, d_2, d_3$} & Vertical distance of the contact point of the first (second or third) phalanx from the proximal connection joint. & mm \\
    
    \multirow{2}{*}{$G_1, G_2, G_3$} & Point of contact of the first (second or third) phalanx with the object. & mm \\
    
    $T$ & Torque transmitted by the actuator to the drive rod. & Nmm \\
    
    $k$ & Coefficient of strength of springs. & N/mm \\
   
    \multirow{2}{*}{$\theta_1, \theta_2, \theta_3$} & Angle between the first, second and third phalanxes and the vertical direction. & rad \\
    \bottomrule
    \end{tabular}
    \end{table}

	In the \textbf{Pinching Mode}, as depicted in Fig \ref{fig:force}a, a coordinate system is established with point P as the origin. To simplify the analysis, only the force on the third phalanx, $F_{3}$, considered as the first and second phalanges exert minimal influence. According to the moment equilibrium condition, the equation is expressed as:
  
    $$\scalebox{0.8}{ \begin{minipage}{\linewidth} \begin{equation}
    \tau=F_{3}\times(d_{3}+L_{2}\cos\theta_{2})    
    \end{equation}\end{minipage} } $$

\begin{figure}[ht]
        \centering
        \includegraphics[width=1\linewidth,keepaspectratio]{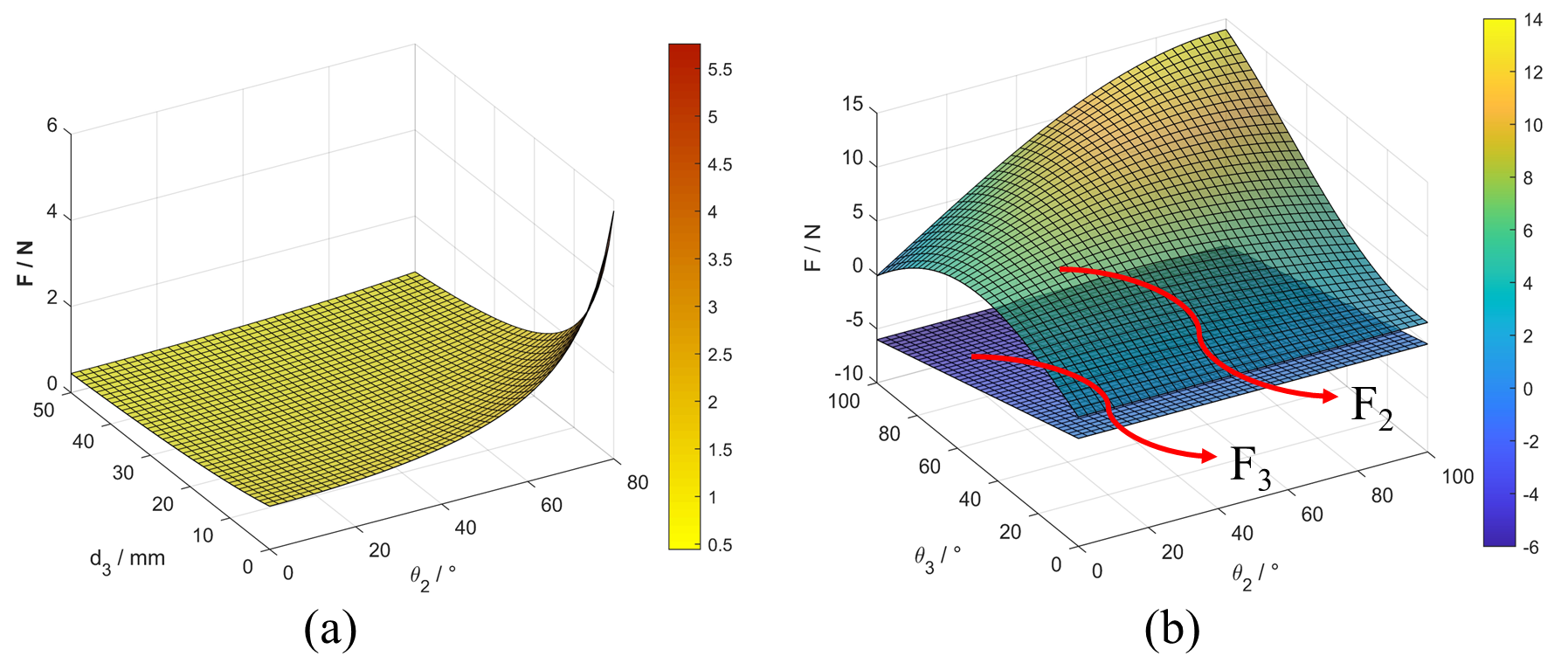}
        \caption{Analysis of the grasp force. (a) Parallel grasp stage;
(b) Self-adaptive grasp stage.}
        \label{fig:MATLAB}
    \end{figure}

	Assuming T=20N$\times$mm and $L_{2}=40$mm, which correspond to the dimensions of the physical prototype, the relationship between the grasping force and the rotation angle of the second finger segment can be determined. This relationship is graphically represented in Fig \ref{fig:MATLAB}a.

	The SPARK Hand transitions into the self-adaptive mode, termed \textbf{Scooping Mode}, when the target object contacts the proximal, intermediate, or distal phalanx impeding their motion. For simplification purposes, this analysis focus on the intermediate and distal phalanges, as the force exerted by the proximal phalanx contributes minimally to the overall grasping force. Fig.\ref{fig:MATLAB}b illustrates the force analysis in the self-adaptive grasping process, wherein the limiting spring applies a counteracting moment.

    During the self-adaptive grasping process, the object primarily interacts with the proximal and intermediate phalanges, while the distal phalanx remains uninvolved in force application or object contact. Consequently, the analysis is confined to the grasping, forces generated by the proximal and intermediate phalanges from the action of scooping the object, the object only interacts with the proximal and intermediate phalanges, while the distal phalange does not exert any force or make contact with the object. Therefore, the analysis is limited to the grasping forces of the proximal and intermediate phalanges.
	
	The force vectors at the contact points between the two phalangeal segments and the object are expressed as:
    $$\scalebox{0.8}{ \begin{minipage}{\linewidth} \begin{equation}
    \vec{F}_{2} = (F_{2}\cos\theta_{2}, -F_{2}\sin\theta_{2}) \label{11}
    \end{equation}\end{minipage} } $$
    $$\scalebox{0.8}{ \begin{minipage}{\linewidth} \begin{equation}
    \vec{F}_{3} = (F_{3}\cos\theta_{3}, -F_{3}\sin\theta_{3})
    \end{equation}\end{minipage} } $$
	The imaginary displacement of the contact point is derived as:
    $$\scalebox{0.8}{ \begin{minipage}{\linewidth} \begin{equation}
    \vec{G}_{1} = (d_{1}\sin\theta_{1}, d_{1}\cos\theta_{1})
    \end{equation}\end{minipage} } $$
    $$\scalebox{0.8}{ \begin{minipage}{\linewidth} \begin{equation}
    \vec{G}_{3} = (L_{2} \sin \theta_{2} + d_{3} \sin \theta_{3}, L_{2} \cos \theta_{2} + d_{3} \cos \theta_{3}) 
    \end{equation}\end{minipage} } $$
	These equations are derived from the Newton-Euler equations.
    $$\scalebox{0.8}{ \begin{minipage}{\linewidth} \begin{equation}
    \begin{bmatrix} T & -k\theta_3 \end{bmatrix} 
    \begin{bmatrix} \delta \theta_2 \\ \delta \theta_3 \end{bmatrix} = 
    \begin{bmatrix} \vec{F}_2 & \vec{F}_3 \end{bmatrix} 
    \begin{bmatrix} \delta\vec{G}_2 \\ \delta\vec{G}_3 \end{bmatrix}
    \end{equation}\end{minipage} } $$
    $$\scalebox{0.8}{ \begin{minipage}{\linewidth} \begin{equation}
    \begin{bmatrix} T & -k\theta_3 \end{bmatrix} = 
    \begin{bmatrix} \vec{F}_2 & \vec{F}_3 \end{bmatrix}  J \label{16}
    \end{equation}\end{minipage} } $$
    $$\scalebox{0.8}{ \begin{minipage}{\linewidth} \begin{equation}
    J = \begin{bmatrix} d_2 & 0 \\ L_2 \cos(\theta_2 - \theta_3) & d_3 \end{bmatrix}
    \end{equation}\end{minipage} } $$
    $$\scalebox{0.8}{ \begin{minipage}{\linewidth} \begin{equation}
    F_2 = \frac{T}{d_2} + \frac{k \theta_3 L_2 \cos(\theta_2 - \theta_3)}{d_2 d_3}, \quad F_3 = -\frac{k \theta_3}{d_3} \label{18}
    \end{equation}\end{minipage} } $$

	In Equation (\ref{16}), \( J \) denotes to the Jacobian matrix. Equations (\ref{11}-\ref{18}) elucidate the relationship between various parameters and the grasping force, as depicted in Fig.\ref{fig:MATLAB}b.

	\section{EXPERIMENTS AND RESULTS}

\begin{figure}[ht]
        \centering
        \includegraphics[width=1\linewidth,keepaspectratio]{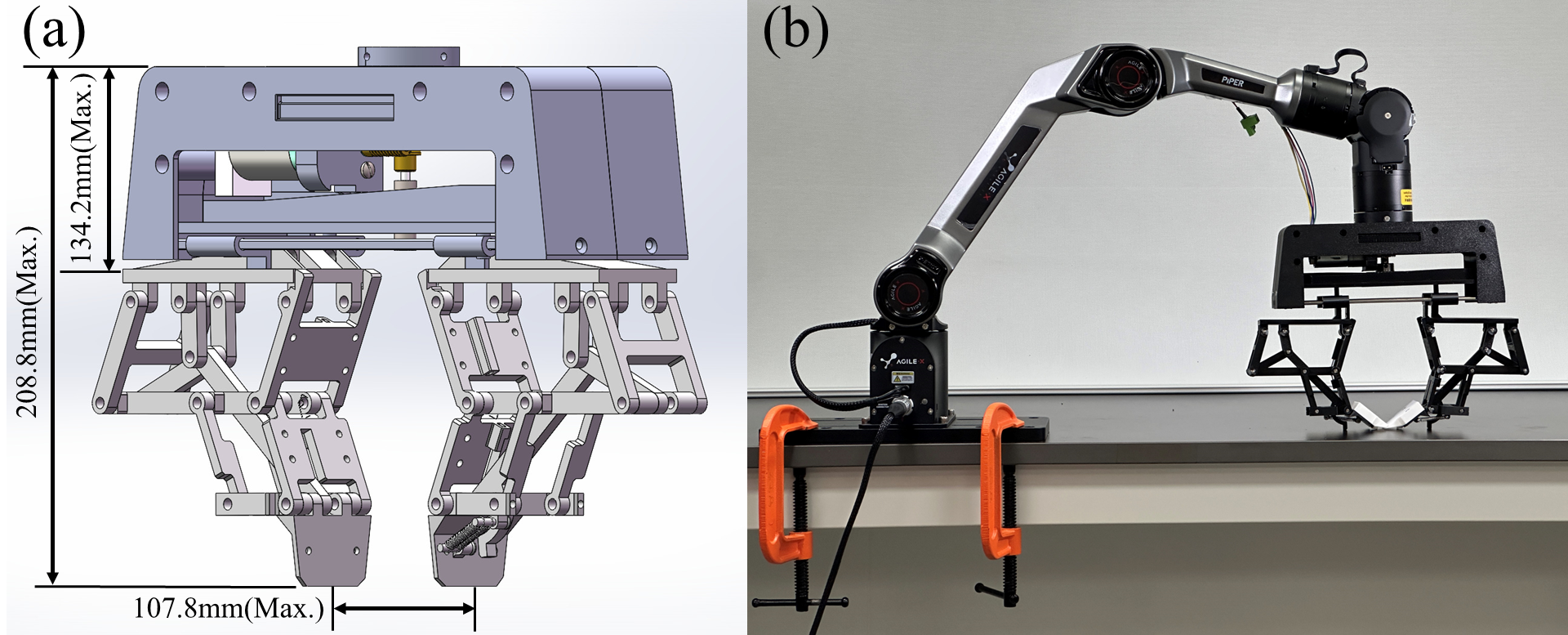}
        \caption{(a) 3D model of the SPARK Hand prototype. (b) Physical implementation of the SPARK Hand mounted on a robotic arm for experimental validation.}
        \label{fig:prototype}
    \end{figure}

    To validate the proposed design's feasibility, a prototype was constructed and subjected to grasping experiments. Fig.~\ref{fig:prototype}a presents the 3D model of the SPARK Hand prototype, while Fig.~\ref{fig:prototype}b showcases the prototype mounted on a robotic arm for experimental validation.

    \begin{figure}[!b]
\centering
		\includegraphics[width=\linewidth]{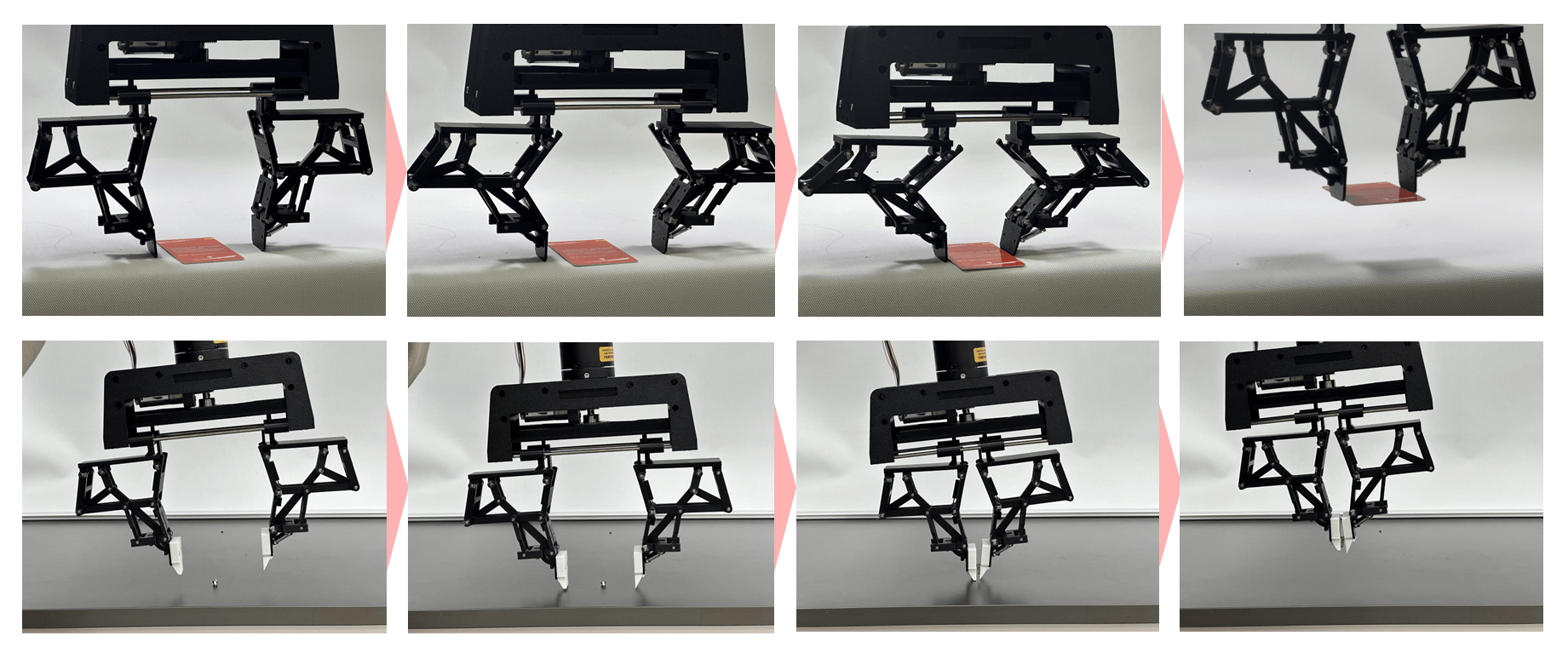}
		\caption{Pinching Grasp: Grasping a thin plastic card and a small screw.}
		\label{fig:Experiment 1}
	\end{figure}
    
    The experiments demonstrated three distinct grasping scenarios: pinching grasp, symmetric scooping grasp, and asymmetric scooping grasp. To enhance scooping performance, white spacers were incorporated into the fingertip design, providing an improved scooping angle.
    
\begin{figure}[h]
		\centering
		\includegraphics[width=\linewidth]{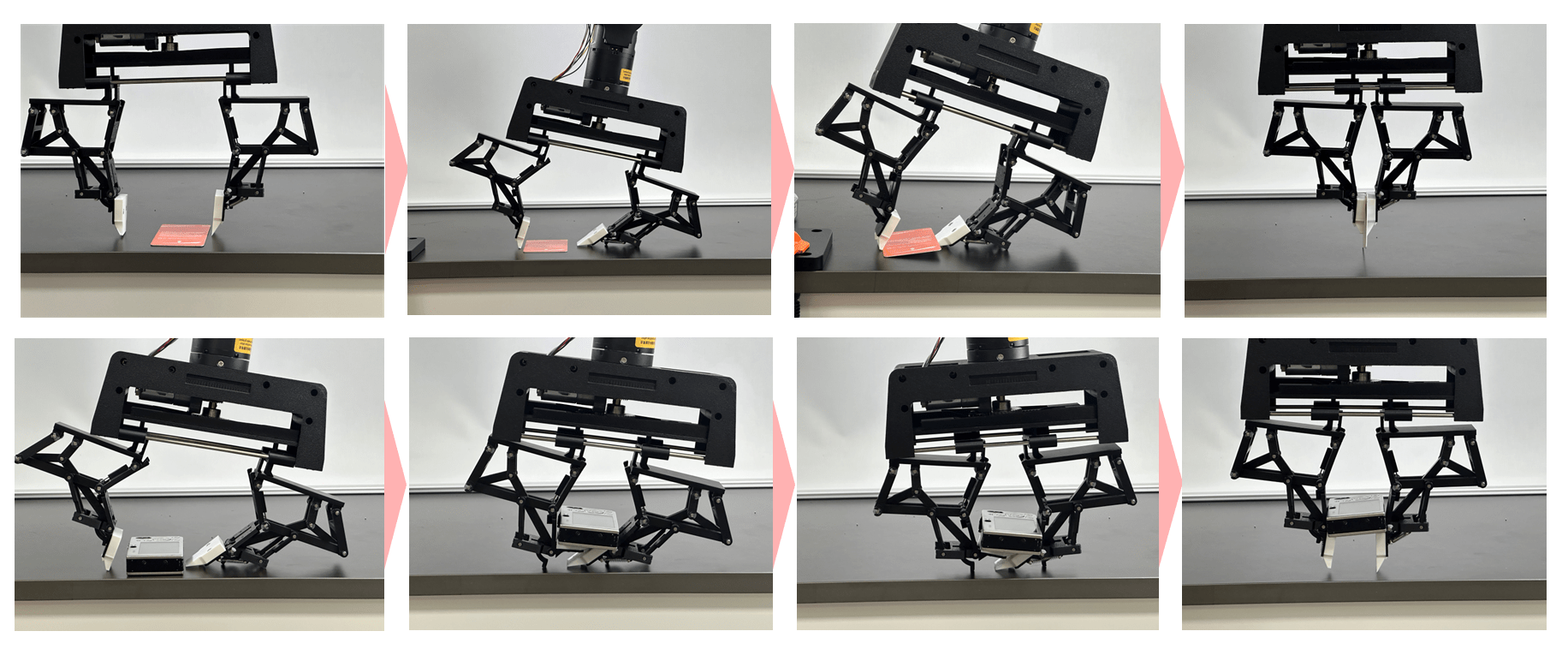}
		\caption{Asymmetric Scooping Grasp by the tilted wrist.Successful pinching grasp with steady object orientation via asymmetric scooping.}
		\label{fig:Experiment 3}
	\end{figure}

	\begin{itemize}
		\item \textbf{Pinching Grasp}: 

    In pinching mode, the robotic fingers execute linear clamping movements to securely grasp objects positioned on a flat surface. The device descends until the distal phalanges contact the surface, after which the fingers close in parallel to clamp the object. Fig.~\ref{fig:Experiment 1} demonstrates the successful grasping of thin plastic cards and small screws, validating the stability and precision of the pinching mode.

		\item \textbf{Symmetric Scooping Grasp}:

        In symmetric scooping mode, the device descends until the distal phalanges contact the surface, subsequently rotating inward to form a scoop-like structure. The two fingers achieve a stable grasp by symmetrically sliding beneath the object. To evaluate this capability, three representative objects—a thin plastic card, a spherical orange, and an irregularly shaped keychain—were tested. Fig.~\ref{fig:Experiment 2} illustrates the symmetric scooping process, demonstrating the device's adaptability in grasping objects of varying shapes and sizes.

		\begin{figure}[h]
		\centering
		\includegraphics[width=\linewidth]{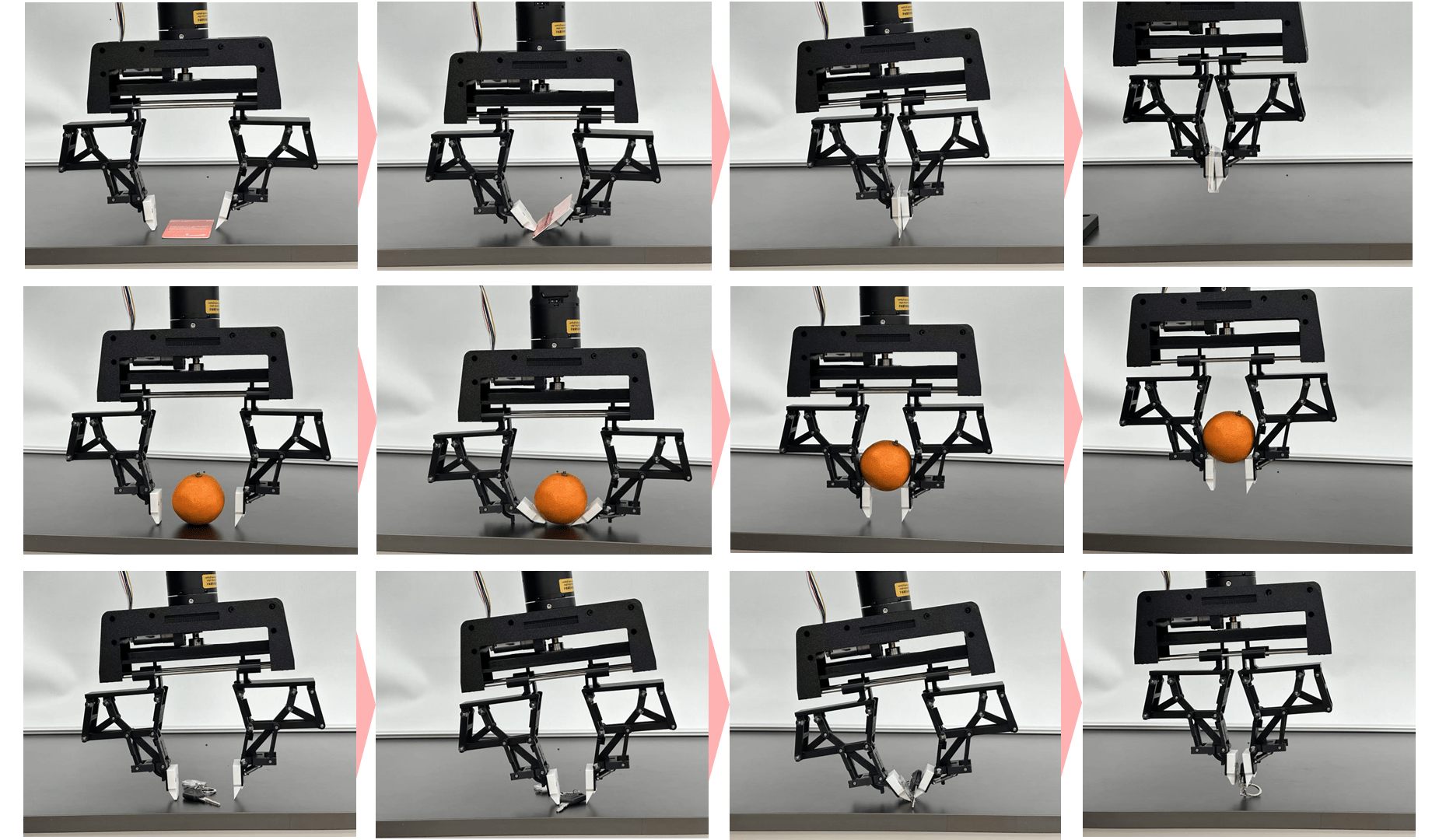}
		\caption{Symmetric Scooping Grasp: Grasping a thin card, orange and keychain.}
		\label{fig:Experiment 2}
	\end{figure}
        
		\item \textbf{Asymmetric Scooping Grasp}: 

        The asymmetric scooping mode was evaluated by tilting the device at a specific angle relative to the table surface. In this configuration, one finger's distal phalanx remains stationary to support the object, while the other finger's distal phalanx rotates inward, forming a shovel-like structure to lift the object from the side. This mode is particularly effective for wide and thin objects, such as flat cards or planar materials.Fig.~\ref{fig:Experiment 3} depicts the asymmetric scooping process, demonstrating the device's capability to adapt to objects with specific geometric characteristics and showcasing its versatility in manipulation tasks.

	\end{itemize}

    In the asymmetric scooping mode, the device is positioned at an oblique angle relative to the table surface. During operation, the distal phalanx of one finger remains stationary, providing support for the object, while the other finger rotates inward to form a shovel-like structure, facilitating lateral lifting. This mode is particularly efficacious for grasping wide and thin objects, such as planar cards or flat materials. Fig.~\ref{fig:Experiment 3} illustrates the asymmetric scooping process, demonstrating the device's adaptability to objects with specific geometric characteristics.

    \section{Conclusion}
    
    This study presents the SPARK finger, an innovative passive adaptive robotic finger mechanism capable of executing hybrid pinching and scooping grasps. By utilizing a multi-link mechanical design incorporating Kempe linkages and elastic elements, the SPARK finger achieves seamless transitions between grasping modes without additional actuators. This capability demonstrates its proficiency in handling a diverse array of object shapes, sizes, and environmental constraints. A prototype system, the SPARK Hand, was developed and experimentally validated, exhibiting its superior performance in grasping thin and flat objects, which typically pose challenges for conventional robotic grippers.
    The SPARK finger's key advantages include dual grasping modes, enabling versatility in handling diverse object geometries, and robust environmental compliance, allowing interaction with surfaces of varying heights and inclinations. Its passive design minimizes control complexity while maintaining stability, adaptability, and precision in constrained environments.

    \bibliographystyle{IEEEtran}
    \bibliography{references}

\begin{thebibliography}{10}
\providecommand{\url}[1]{#1}
\csname url@samestyle\endcsname
\providecommand{\newblock}{\relax}
\providecommand{\bibinfo}[2]{#2}
\providecommand{\BIBentrySTDinterwordspacing}{\spaceskip=0pt\relax}
\providecommand{\BIBentryALTinterwordstretchfactor}{4}
\providecommand{\BIBentryALTinterwordspacing}{\spaceskip=\fontdimen2\font plus
\BIBentryALTinterwordstretchfactor\fontdimen3\font minus \fontdimen4\font\relax}
\providecommand{\BIBforeignlanguage}[2]{{%
\expandafter\ifx\csname l@#1\endcsname\relax
\typeout{** WARNING: IEEEtran.bst: No hyphenation pattern has been}%
\typeout{** loaded for the language `#1'. Using the pattern for}%
\typeout{** the default language instead.}%
\else
\language=\csname l@#1\endcsname
\fi
#2}}
\providecommand{\BIBdecl}{\relax}
\BIBdecl

\bibitem{salisbury1982articulated}
J.~K. Salisbury and J.~J. Craig, ``Articulated hands: Force control and kinematic issues,'' \emph{Int. J. Robot. Res.}, vol.~1, no.~1, pp. 4--17, 1982.

\bibitem{jacobsen1986design}
S.~Jacobsen, E.~Iversen, and D.~Knutti, ``Design of the utah/mit dexterous hand,'' in \emph{IEEE Int. Conf. on Robotics and Automation (ICRA)}.\hskip 1em plus 0.5em minus 0.4em\relax IEEE, 1986, pp. 1520--1532.

\bibitem{lovchik1999robonaut}
C.~S. Lovchik and M.~Diftler, ``The robonaut hand: A dexterous robot hand for space,'' in \emph{IEEE Int. Conf. on Robotics and Automation (ICRA)}.\hskip 1em plus 0.5em minus 0.4em\relax IEEE, 1999, pp. 907--912.

\bibitem{wei2005fpga}
R.~Wei, X.~H. Gao, and M.~H. Jin, ``Fpga based hardware architecture for hit/dlr hand,'' in \emph{IEEE Int. Conf. on Intelligent Robots and Systems (IROS)}.\hskip 1em plus 0.5em minus 0.4em\relax IEEE, 2005, pp. 3233--3238.

\bibitem{brown2010universal}
E.~Brown, N.~Rodenberg, J.~Amend \emph{et~al.}, ``Universal robotic gripper based on the jamming of granular material,'' \emph{Proceedings of the National Academy of Sciences}, vol. 107, no.~44, pp. 18\,809--18\,814, 2010.

\bibitem{bauer2014soft}
S.~Bauer, S.~Bauer-Gogonea, I.~Graz \emph{et~al.}, ``25th anniversary article: A soft future: From robots and sensor skin to energy harvesters,'' \emph{Advanced Materials}, vol.~26, no.~1, pp. 149--162, 2014.

\bibitem{laliberte2002underactuation}
T.~Laliberté, L.~Birglen, and C.~Gosselin, ``Underactuation in robotic grasping hands,'' \emph{Machine Intelligence and Robotic Control}, vol.~4, no.~3, pp. 1--11, 2002.

\bibitem{townsend2000barrett}
W.~Townsend, ``The barretthand grasper – programmably flexible part handling and assembly,'' \emph{Ind. Robot}, vol.~27, no.~3, pp. 181--188, 2000.

\bibitem{deimel2016novel}
R.~Deimel and O.~Brock, ``A novel type of compliant and underactuated robotic hand for dexterous grasping,'' \emph{Int. J. Robot. Res.}, vol.~35, no. 1-3, pp. 161--185, 2016.

\bibitem{chen2024novel}
S.~Chen \emph{et~al.}, ``A novel geometrical structure robot hand for linear-parallel pinching and coupled self-adaptive hybrid grasping,'' in \emph{IEEE/RSJ Int. Conf. on Intelligent Robots and Systems (IROS)}, 2024, pp. 3030--3035.

\bibitem{yoon2022fullypassive}
D.~Yoon and K.~Kim, ``Fully passive robotic finger for human-inspired adaptive grasping in environmental constraints,'' \emph{IEEE/ASME Trans. Mechatronics}, vol.~27, pp. 3841--3852, 2022.

\bibitem{kempe1877}
A.~B. Kempe, ``How to draw a straight-line,'' \emph{Nature}, vol.~16, no. 877, pp. 65--67, 86--89, 125--127, and 145--146, 1877.

\bibitem{dijksman1975}
E.~A. Dijksman, ``Kempe's linkages and their derivations,'' \emph{J. Eng. Ind.}, vol.~97, no. August, pp. 801--806, 1975.

\end{thebibliography}
	
\end{document}